\documentclass[letterpaper]{article} 

\usepackage{paperstyle}
\nocopyright   

\usepackage[hyphens]{url}  
\usepackage{graphicx}      
\usepackage{natbib}        
\usepackage{caption}       
\usepackage{amsmath}
\usepackage{amssymb}
\usepackage{booktabs}

\title{Mixture of Channel Experts: Static Sparse Supports with\\Input-Adaptive Mixing for Pointwise Projections}

\author{
Elian Iluk,
Gil Ben-Artzi
}

\affiliations{
School of Computer Science, Ariel University, Israel\\
elianroy.iluk@msmail.ariel.ac.il,
gilba@ariel.ac.il
}

\begin{document}
\maketitle

\begin{abstract}
Mixture-of-Experts (MoE) scales language models by routing each input through a small set of independently parameterized experts. We show that copying this design into convolutional networks fails for a structural reason: parallel convolutional experts that read the same input channels learn nearly identical filters. We therefore move the expert axis from \emph{operator duplication} to \emph{channel selection}. We introduce \emph{Mixture of Channel Experts (MoCE)}, a structured sparse channel-mixing layer, inspired by MoE, that replaces pointwise ($1\times1$) channel-reduction projections. In MoCE, an \emph{expert} is a single output channel with a learned sparse support of $k \ll C$ input channels. The selected channels are combined by a softmax whose temperature is predicted per input, so each expert can move between mean-like and max-like aggregation. A residual expert summarizes the unselected channels, and a load-balancing loss keeps channel coverage complete.  MoCE replaces a dense projection whose cost is quadratic in $C$ with a mechanism whose relative cost scales as $k/C$, and the predicted savings hold in measured wall-clock time. Across ResNet backbones on ImageNet-1K and CIFAR-100, transfer learning, EfficientViT, and a strong modern training recipe, MoCE matches or exceeds dense baselines and prior channel-selection methods while reducing MACs by 16.7\% and end-to-end latency.
\end{abstract}

\section{Introduction}
\label{sec:intro}

Pointwise projections mix channels independently at every spatial location and appear throughout modern vision backbones: as the entrance and exit stages of ResNet bottlenecks, as the projections of inverted bottlenecks and ConvNeXt-style blocks, and---since a token-wise linear map is a $1\times1$ convolution---as the feed-forward projections of vision transformers. Their cost is $C_{\mathrm{in}}C_{\mathrm{out}}HW$, so channel width directly increases both arithmetic and parameter count. This makes them a natural target for structured sparsification, provided the resulting operator remains efficient on hardware.

\begin{figure}[t]
\centering
\includegraphics[width=0.76\linewidth]{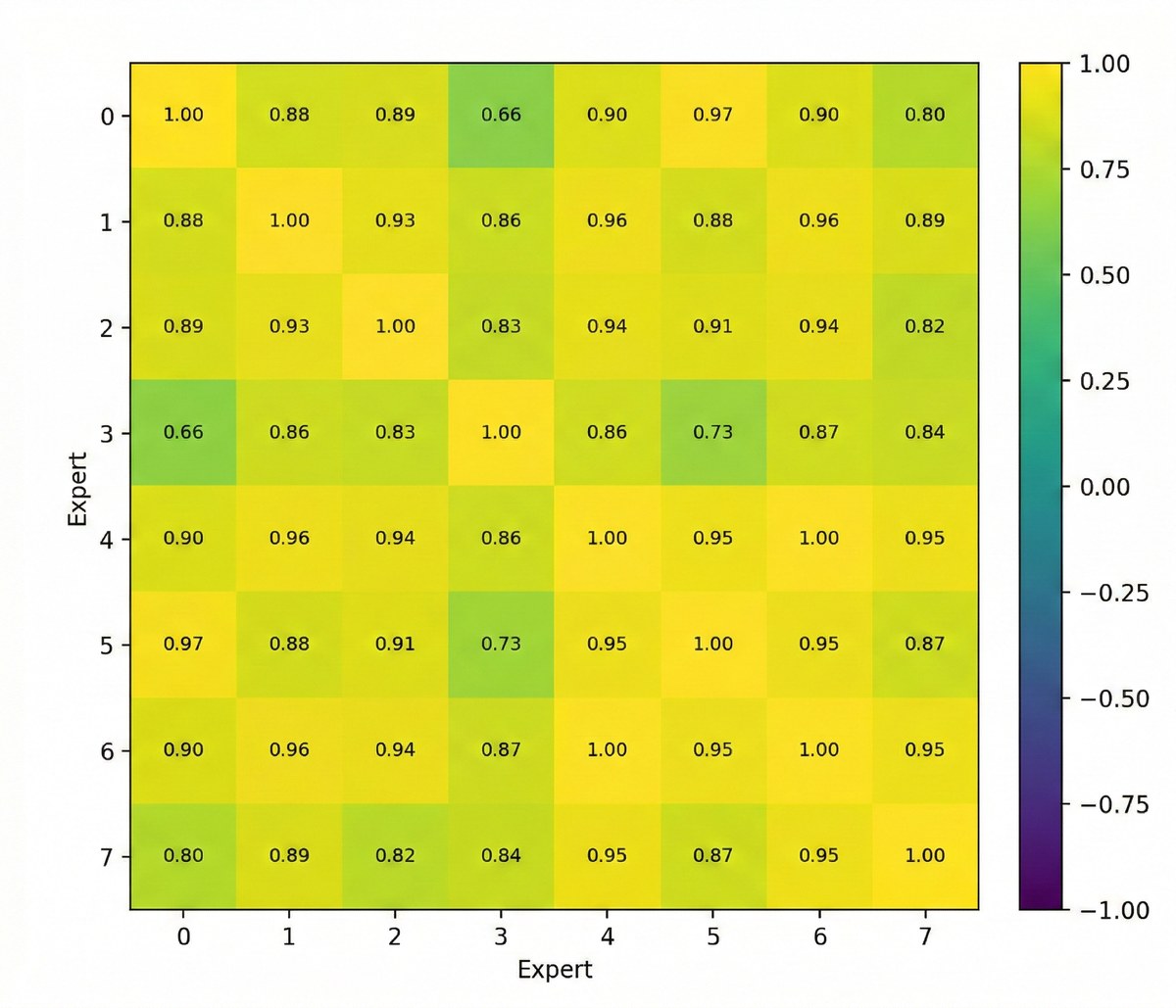}
\caption{\textbf{Parallel convolutional experts sharing an input} (ResNet-50, CIFAR-100): pairwise kernel cosine similarity.}
\label{fig:conv_similarity}
\end{figure}

\begin{figure}[t]
\centering
\includegraphics[width=\linewidth]{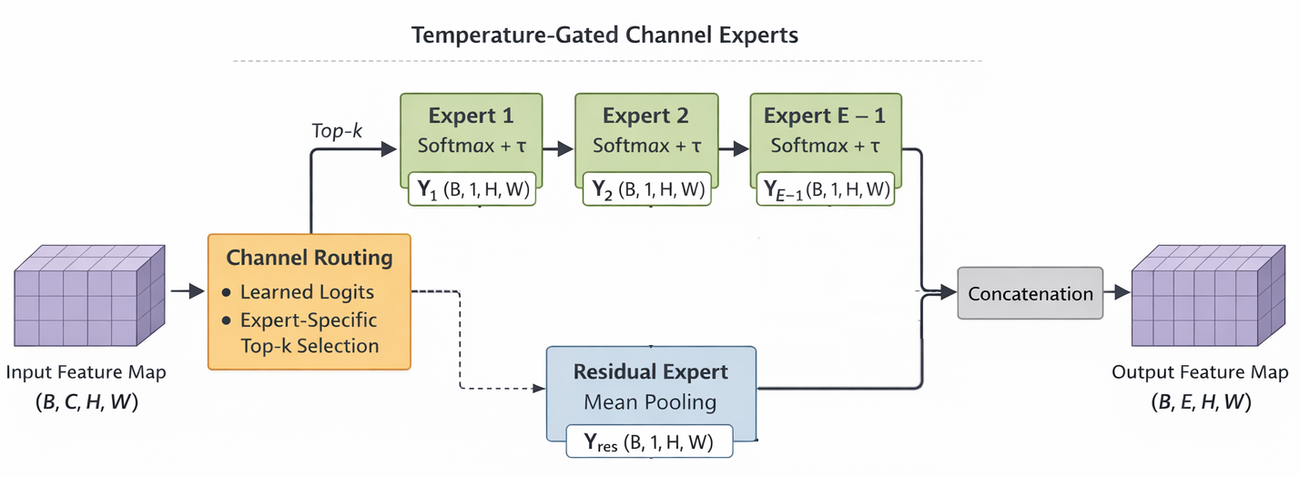}
\caption{\textbf{MoCE layer.} $E\!-\!1$ routed experts over learned static supports, plus one residual aggregate.}
\label{fig:overview}
\end{figure}

A direct convolutional analogue of Mixture-of-Experts (MoE) duplicates an operator and routes each input to a subset of parallel experts. In a controlled ResNet-50/CIFAR-100 diagnostic with eight convolutional experts sharing the same input representation and objective, the learned kernels are strongly aligned: the mean off-diagonal cosine similarity is $0.88$, with most pairs between $0.85$ and $0.97$ (Fig.~\ref{fig:conv_similarity}). A comparable collapse has been reported for upcycled transformer MoE, where experts remain close to the original dense weights and to one another~\citep{Huang2025DeRS}. Together these observations indicate that operator duplication can spend parameters without producing useful specialization when experts receive the same representation and supervision. We therefore move specialization from duplicated operators to sparse channel supports. \emph{Mixture of Channel Experts} (MoCE) replaces a dense pointwise projection with one aggregation unit per output channel (Fig.~\ref{fig:overview}). Each routed expert selects a learned top-$k$ subset of the input channels and forms a convex mixture over that support. The support is static at inference, enabling fixed gathers and packed execution. A lightweight gate predicts one temperature per expert and example; the temperature changes how concentrated the expert's learned channel preferences are, without changing the selected support or the ordering of its weights. One residual output aggregates channels not selected by any routed expert.

\paragraph{Where input dependence pays.}
The design question that organizes this paper is whether the input signal is better spent on choosing \emph{which} channels an expert reads or on choosing \emph{how} it combines them. The dynamic alternative is the more expressive of the two: a router must score every candidate channel before selecting, so it reads the full pooled descriptor and adds a low-rank offset to the routing logits, which changes both which channels are read and how they are weighted. The temperature gate, by contrast, reads only the $k$ descriptors already on the expert's support and emits a single scalar. Despite this advantage, we show that for example, on CIFAR-100 the dynamic variant improves the accuracy by $+0.03$ points while making the routing-and-gather path $7.13\times$ slower, because the memory-access pattern becomes data dependent. However, removing the temperature gate reduces accuracy by $1.19$ points. That the more expressive route produces no meaningful gain in the evaluated run, while removing the more constrained one costs over a point, is what makes static supports the better operating point, and it is why MoCE remains statically schedulable.

MoCE is inspired by expert-side routing, but it is not a capacity-scaling MoE: an expert is an output channel rather than an independently parameterized network. Its closest methodological context is structured sparse projection, channel routing, dynamic channel selection, and channel attention. The contribution is the combination of learned static support, minimal input-adaptive mixing, residual coverage, and an implementation whose savings are visible in wall-clock measurements. Across ResNet and EfficientViT backbones, sparse supports preserve or improve the dense accuracy--efficiency operating point at $17$--$21\%$ fewer MACs, with deployed-parameter reductions of $17$--$21\%$ on the ResNet variants, and an end-to-end speedup that is smaller than the MAC reduction for reasons an arithmetic-intensity analysis makes explicit.

\section{Related Work}

\paragraph{Sparse experts and routing.}
Sparse MoE activates a subset of experts for each input, making routing balance and stability central concerns~\citep{Shazeer2017,Lepikhin2021GShard,Fedus2021Switch}. Expert-Choice routing reverses token-to-expert assignment and gives experts explicit capacity~\citep{Zhou2022ExpertChoice}; differentiable and soft alternatives reduce discontinuities in hard assignment~\citep{Hazimeh2021DSelectK,Puigcerver2024SoftMoE}. Vision MoE primarily applies these ideas to token routing in transformers~\citep{Riquelme2021VMoE,Liu2024RoutersMoE}. MoCE borrows expert-side support selection and aggregate balancing, but applies them along the channel axis of a pointwise projection: channels play the role that tokens play in MoE, and each expert selects its own top-$k$ of them. Token features vary with every input and are therefore ordinarily routed dynamically; whether channel membership benefits similarly from dynamic routing is what our membership experiment tests.

\paragraph{Structured sparse pointwise projection.}
Structured sparsity learning removes groups of weights during training~\citep{Wen2016SSL}. CondenseNet learns group-sparse $1\times1$ connectivity and freezes it for inference~\citep{Huang2018CondenseNet}; movement-based and wiring-based methods also learn which connections survive optimization~\citep{Sanh2020Movement,Wortsman2019DNW}. MoCE shares the static-support principle but constrains each routed output to normalized nonnegative weights and retains one scalar of input dependence that modulates their concentration. It should therefore be read as a sparse projection with an adaptive mixing rule, rather than as a replacement for the broader structured-pruning literature.

\paragraph{Conditional computation in CNNs.}
Channel gating, dynamic pruning, and feature suppression select or reweight channels per input~\citep{Hua2019ChannelGating,Gao2019FBS,Bejnordi2020BatchShaping,Gao2024BilevelPruning}. Dynamic group connectivity and joint spatial--channel gating change the active operator structure~\citep{Su2020DGC,Li2021DGNet}. Pick-or-Mix performs fine-grained per-input channel sampling~\citep{Kumar2024PickOrMix}, while Squeeze-and-Excitation recalibrates channels from global descriptors~\citep{Hu2018SENet}. CondConv and Dynamic Convolution form input-dependent mixtures of dense kernels~\citep{Yang2019CondConv,Chen2020DynamicConv}. MoCE instead removes channel connectivity and keeps the data-dependent component scalar and support preserving.

\section{Mixture of Channel Experts}
\label{sec:method}

\subsection{Setting and Scope}

Let $X \in \mathbb{R}^{B\times C\times H\times W}$ and consider a pointwise map from $C$ input channels to $E \le C$ outputs. A dense layer computes
\begin{equation}
Y_o(h,w) = \sum_{c=1}^{C} W_{o,c} X_c(h,w)
\end{equation}
with $CEHW$ MACs. MoCE emits the same shape using $E-1$ routed experts and one residual aggregate.

MoCE applies to any pointwise projection with $E \le C$; the ratio $s = C/E$ is a property of the layer, not of the method. In ResNet we replace the bottleneck entrance projection, where $s$ is $4$ for most layers and $1$ or $2$ at stage boundaries. In EfficientViT we replace the second feed-forward projection of each ConvMlp block, where $s = 2$. The method does not replace spatial convolution.

\subsection{Learning Static Supports}

Routing logits $L \in \mathbb{R}^{(E-1)\times C}$ assign one preference vector to each routed expert. Its support is
\begin{equation}
\mathcal{S}_e = \operatorname{TopK}(L_e, k).
\label{eq:support}
\end{equation}
The forward pass uses this hard support. The task loss differentiates through the selected logit values used by the mixer, while the full-support coverage loss in Sec.~\ref{sec:coverage} differentiates through every logit. No gradient is assigned through the discrete indices themselves; supports change when the continuous logits reorder during training. At inference, each $\mathcal{S}_e$ is precomputed and fixed, so the layer requires neither top-$k$ search nor data-dependent branching, and its memory-access pattern can be scheduled ahead of time.

\subsection{Input-Adaptive Mixing}
\label{sec:temperature}

Global average pooling gives $z = \operatorname{GAP}(X) \in \mathbb{R}^{B\times C}$. A two-layer gate $g:\mathbb{R}^{k}\rightarrow\mathbb{R}$, shared by the experts within a layer, receives the $k$ descriptors on the expert's own support and predicts
\begin{equation}
\tau_e(X) = \tau_{\min} + (\tau_{\max}-\tau_{\min})\,\sigma\!\left(g(z_{\mathcal{S}_e})\right).
\end{equation}
Because $g$ is not permutation invariant, the ordering of its input is part of the specification: the $k$ descriptors are supplied in order of descending routing logit, so position $i$ of the gate input always corresponds to the expert's $i$-th most preferred channel. This shared positional meaning is what makes a single gate sensible across experts, and the same rule is applied to the example-dependent support of Sec.~\ref{sec:adaptivity}.
The selected logits define mixing coefficients
\begin{equation}
a_{e,i}(X) = \frac{\exp\!\big(L_{e,i}/\tau_e(X)\big)}{\sum_{j\in\mathcal{S}_e}\exp\!\big(L_{e,j}/\tau_e(X)\big)},
\quad i \in \mathcal{S}_e,
\label{eq:weights}
\end{equation}
and the routed output is
\begin{equation}
Y_e(h,w) = \sum_{i\in\mathcal{S}_e} a_{e,i}(X)\, X_i(h,w).
\label{eq:expert}
\end{equation}
The weights are shared over spatial locations and vary across examples only through $\tau_e(X)$. Small temperatures concentrate mass on the selected channel with the largest routing logit; large temperatures approach a uniform average over the support. The temperature therefore changes the entropy of fixed routing preferences, not the maximum activation value. For a fixed $L_e$, the reachable weight vectors trace a one-dimensional curve on the $(k\!-\!1)$-simplex along which the ordering of the weights is invariant: per expert and example, the adaptive capacity of the layer is exactly one scalar, which is why it is nearly free to compute.

For any positive constant $\tau$, $\operatorname{softmax}(L_e/\tau)$ can be represented as $\operatorname{softmax}(L'_e)$ with $L'_e = L_e/\tau$, and positive rescaling leaves top-$k$ unchanged. Consequently, the $\tau\equiv1$ ablation has the same static representational class as any model with a learned input-independent temperature. Optimization and interaction with the coverage term can still differ, so we use this equivalence to define a function-class-matched static control rather than to claim a causal proof. The measured gap nevertheless provides direct evidence that example-dependent concentration is useful beyond static sparse mixing. Two further properties keep the temperature from being redundant in the full model: it depends on the input through $z_{\mathcal{S}_e}$, which no reparameterization of a static $L$ can express; and it divides $L$ in Eq.~\ref{eq:weights} but not in the coverage objective, so it decouples how sharply an expert mixes from how sharply the regularizer perceives it to mix.

\begin{figure}[t]
\centering
\includegraphics[width=0.72\linewidth]{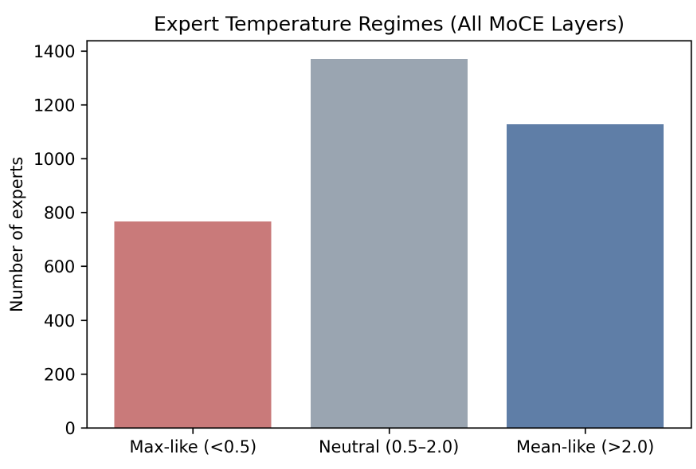}
\caption{\textbf{Learned temperatures across experts,} spanning concentrated, intermediate, and near-uniform regimes rather than collapsing to one mode.}
\label{fig:temp_regimes}
\end{figure}

Equation~\ref{eq:expert} follows the softmax-mixing principle of attention, with the weights derived from learned routing preferences rather than query--key interactions and computed over a sparse static support. It is also a convex mixture, and therefore strictly less expressive than an unconstrained signed projection on the same support; we treat this as a structured regularizer with an execution benefit, and Sec.~\ref{sec:adaptivity} reports how much of the dense projection's function is actually required. Routing behavior is visualized in Figure~\ref{fig:routing}. Figure~\ref{fig:temp_regimes} shows the distribution of learned temperatures over all MoCE experts.

\subsection{Residual Coverage and Regularization}
\label{sec:coverage}

Let $\mathcal{U} = \{1,\ldots,C\}\setminus\bigcup_e \mathcal{S}_e$. The residual output is
\begin{equation}
Y_{\mathrm{res}}(h,w) =
\begin{cases}
|\mathcal{U}|^{-1}\sum_{i\in\mathcal{U}} X_i(h,w), & |\mathcal{U}|>0,\\
0, & |\mathcal{U}|=0.
\end{cases}
\end{equation}
Thus every input channel contributes either through a routed expert or through a compressed residual summary. On average, $7.7\%$ of input channels remain uncovered across the replaced layers. Replacing the mean with a temperature-gated aggregation over $\mathcal{U}$ changed accuracy negligibly, so we keep the cheaper form.

\begin{figure}[t]
\centering
\includegraphics[width=\linewidth]{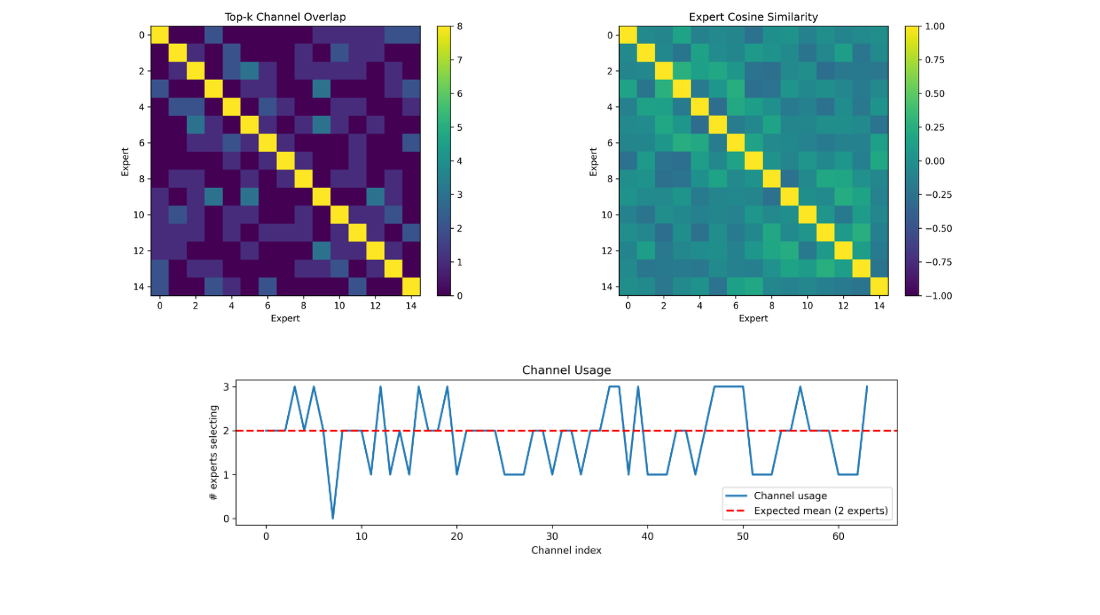}
\caption{\textbf{Routing in the first MoCE layer.} Top left: top-$k$ overlap between experts. Top right: cosine similarity of routing-logit rows. Bottom: experts selecting each channel, against the uniform expectation.}
\label{fig:routing}
\end{figure}

To discourage concentration on a small channel subset, define $p_{e,c} = \operatorname{Softmax}(L_e)_c$ and $u_c = (E-1)^{-1}\sum_e p_{e,c}$. We optimize
\begin{equation}
\mathcal{L}_{\mathrm{cov}} = \frac{\sum_c (u_c-\bar{u})^2}{\bar{u}^2+\epsilon},
\qquad \bar{u} = C^{-1}\sum_c u_c,
\end{equation}
with $\mathcal{L} = \mathcal{L}_{\mathrm{task}} + \lambda\mathcal{L}_{\mathrm{cov}}$. This regularizer equalizes aggregate soft usage; it does not guarantee disjoint hard supports, and Fig.~\ref{fig:routing} reports the resulting overlap and usage empirically. Its contribution to coverage alone should not be overstated: with $(E-1)k$ selections spread over $C$ channels, even independent uniform supports would leave only a modest fraction uncovered, so the regularizer improves an already-high baseline. It also supplies gradients to unselected logits and thereby permits their ordering, and hence the hard supports, to change during training.

\subsection{Cost, Break-Even, and Bandwidth}
\label{sec:cost}

The dominant per-layer spatial work is
\begin{equation}
M_{\mathrm{MoCE}} = (E-1)kHW + CHW + |\mathcal{U}|HW + M_g,
\label{eq:cost}
\end{equation}
where the first term is routed mixing, the second is global pooling, the third is the residual reduction, and $M_g$ is the small non-spatial gate. Relative to $M_{\mathrm{dense}} = CEHW$, the leading term is approximately $k/C$; with $|\mathcal{U}| \le C$, a conservative upper bound on the ratio is
\begin{equation}
\frac{M_{\mathrm{MoCE}}}{M_{\mathrm{dense}}} \lesssim \frac{k}{C} + \frac{2}{E}.
\label{eq:ratio}
\end{equation}
The residual and pooling overheads scale only linearly in $C$ while the dense projection scales as $CE$, so the advantage grows with input width. Using the measured $|\mathcal{U}|\approx0.077C$ instead of the worst case tightens the second term to $\approx 1.08/E$; all reported MoCE MAC values use the conservative bound.

\paragraph{Break-even.} Parity requires $(E-1)k + 2C = CE$, giving
\begin{equation}
k^\star = \frac{C(E-2)}{E-1}.
\end{equation}
For the common $s=4$ projections this is $k^\star = 252$ at $C=256$ and $k^\star = 2044$ at $C=2048$. The threshold is smallest at the $64\rightarrow64$ stage-entry projection, where $k^\star = 63$, and exceeds $250$ for every other replaced layer. At $k=8$ the layer therefore operates well below parity everywhere, which means that within this design the ceiling on $k$ is representational rather than computational---a point the sparsity sweep in Sec.~\ref{sec:sweeps} confirms from the other direction.

\paragraph{Instantiation.} On ResNet-50 the sixteen replaced projections account for $706.5$\,M spatial MACs, $17.2\%$ of the network budget, and MoCE reduces them to $20.1$\,M: a $35\times$ reduction on the replaced operators and $16.7\%$ of the whole network (Table~\ref{tab:cost}). Per-layer costs follow directly from Eq.~\ref{eq:cost}; the omitted non-spatial gate contributes $0.00027$\,G and does not affect the rounded totals. The saving grows to $20.8\%$ on ResNet-152, since deeper backbones place proportionally more replaced projections in wide stages.

\paragraph{Storage.} Training stores the full routing matrix, $(E-1)C$ values per layer, so that supports can migrate; at that point MoCE is not a parameter reduction, and its training-time count sits within $0.05$\,M of the dense model. Deployment retains only the $(E-1)k$ selected logits, their integer indices, and the shared gate. In ResNet-50, the $4.33$\,M weights of the replaced dense projections are replaced at deployment by approximately $30.1$\,k retained logits and $60$\,KB of 16-bit support indices, reducing the complete model from $25.56$\,M to $21.26$\,M learned parameter values; the corresponding reduction is $19.6\%$ on ResNet-101 and $20.5\%$ on ResNet-152. We report both training and deployment counts throughout, and the index overhead wherever total storage is at issue.

\paragraph{Why the time saving is smaller than the MAC saving.} Both the dense projection and MoCE move $(C+E)HW$ activation elements per example. Sparsification removes arithmetic but not traffic: the pooling reads all $C$ channels and the residual reads $|\mathcal{U}|$ of them, so activation traffic remains $\Theta(CHW)$ independently of $k$. A roofline-style estimate from these idealized traffic counts, at the batch size $512$ used in Sec.~\ref{sec:latency} where weight traffic is heavily amortized, places the dense $2048\rightarrow512$ projection near $205$ FLOP/byte in FP32 and MoCE near $1.6$ FLOP/byte, against an accelerator balance point around $13$ FLOP/byte. This suggests that MoCE shifts the projection from a compute-dominated toward a bandwidth-dominated regime, which would explain why a $128\times$ MAC reduction on the deepest replaced projection yields a $3.0\times$ measured speedup. We present this as an explanatory estimate rather than a profiler-verified attribution; its practical implication is that reducing activation traffic---fusing the pooling into the gather, and evaluating the gate as a single batched product---offers more headroom than further reductions in $k$.

\begin{table}[t]
\centering
\scriptsize
\setlength{\tabcolsep}{3pt}
\caption{Cost and storage, accounted as in Sec.~\ref{sec:cost}: MACs are network totals with each replaced projection counted as $[(E{-}1)k + 2C]HW$ under the conservative bound $|\mathcal{U}|{=}C$. MoCE parameters are training/deployment counts of learned values.}
\label{tab:cost}
\begin{tabular}{lccccc}
\toprule
& \multicolumn{2}{c}{MACs (G)} & & \multicolumn{2}{c}{Params (M)} \\
\cmidrule(lr){2-3}\cmidrule(lr){5-6}
Backbone & Dense & MoCE & & Dense & MoCE (tr./dep.) \\
\midrule
ResNet-50  & 4.112  & \textbf{3.426} & & 25.56 & 25.55 / \textbf{21.26} \\
ResNet-101 & 7.834  & \textbf{6.288} & & 44.55 & 44.52 / \textbf{35.83} \\
ResNet-152 & 11.559 & \textbf{9.156} & & 60.19 & 60.15 / \textbf{47.84} \\
\bottomrule
\end{tabular}
\end{table}

\section{Experiments}
\label{sec:experiments}

\subsection{Setup}

We evaluate ImageNet-1K~\citep{Deng2009ImageNet}, CIFAR-100~\citep{krizhevsky2009learning}, ImageNet-to-CIFAR transfer, and EfficientViT~\citep{Liu2023EfficientViT}. ImageNet models train for $120$ epochs with batch size $512$, SGD with momentum $0.9$ and weight decay $10^{-4}$, base learning rate $0.1$ with cosine decay, and standard random-resized-crop to $224^2$ with horizontal flipping; no label smoothing, mixup, or CutMix is used. Transfer fine-tuning uses $20$ epochs at learning rate $0.05$; during fine-tuning the support indices are frozen while the selected logits, temperature gates, and all remaining network parameters are updated. We deliberately adopt this original-style ResNet recipe rather than a high-accuracy one: it matches the protocol under which the compared methods were developed, and it isolates the architectural change from the training pipeline. Our $75.98\%$ ResNet-50 baseline is consistent with standard original-recipe implementations.

Accuracies in the main tables are means and standard deviations over three seeds; ablation entries are single runs and are described as such throughout. With three runs we make no claim of statistical significance; we interpret differences that are small relative to the observed run-to-run variation as accuracy preservation rather than as improvement. Bold typeface is used only for deterministic quantities (MACs and parameter counts), never for accuracies.

All ResNet bottleneck entrance projections are replaced ($16$, $33$, and $50$ layers for ResNet-50/101/152), and BatchNorm and the surrounding block structure are unchanged. Unless stated otherwise, $k=8$, $\tau\in[0.2,6.0]$, and $\lambda = 5\times10^{-4}$ on ImageNet and $0.05$ on CIFAR-100. Routing logits are initialized i.i.d.\ from a zero-mean Gaussian, and the temperature gate is a two-layer MLP of hidden width $k$ with ReLU and biases, shared across the experts of a layer. EfficientViT uses $k=64$ and replaces the second $2d\rightarrow d$ feed-forward projection in each ConvMlp block, in both \texttt{ffn0} and \texttt{ffn1} across all stages, leaving expansion, activation, and attention modules unchanged. CIFAR-100 models trained from scratch use a CIFAR-style stem, a single $3\times3$ convolution with no max-pooling, giving stage resolutions of $32/16/8/4$. The supplement gives the complete training recipes for the CIFAR-from-scratch and EfficientViT settings---epochs, batch size, optimizer, learning-rate schedule, warm-up, weight decay, and augmentation---together with the routing-logit initialization variance, the Gumbel temperature schedule used for the dynamic ablation, the warm-up and timed-iteration counts for the latency measurements, and the GPU model with framework, CUDA, and cuDNN versions.

\subsection{ImageNet-1K}

Table~\ref{tab:imagenet} reports matched dense and MoCE models. On ResNet-50 the Top-1 point estimate increases by $0.73$ points while MACs fall by $16.7\%$ and deployed learned parameters by $16.8\%$. On ResNet-101 accuracy is preserved within the observed run-to-run variation at $19.7\%$ lower MACs, and on ResNet-152 the point estimate increases by $0.46$ points at $20.8\%$ lower MACs. Because MoCE improves or preserves accuracy while reducing computation and deployment storage simultaneously, the relevant comparison is the joint operating point rather than the accuracy delta alone.

\begin{table}[t]
\centering
\scriptsize
\setlength{\tabcolsep}{3.5pt}
\caption{ImageNet-1K; accuracies are mean $\pm$ standard deviation over three seeds. MoCE parameters are training/deployment counts of learned values; deployment additionally stores packed support indices ($\approx60$\,KB for ResNet-50).}
\label{tab:imagenet}
\begin{tabular}{lccc}
\toprule
Model & Params (M) & MACs (G) & Top-1 (\%) \\
\midrule
ResNet-50 & 25.56 & 4.112 & 75.98 $\pm$ 0.30 \\
\quad + MoCE & 25.55 / \textbf{21.26} & \textbf{3.426} & 76.71 $\pm$ 0.20 \\
\midrule
ResNet-101 & 44.55 & 7.834 & 77.21 $\pm$ 0.29 \\
\quad + MoCE & 44.52 / \textbf{35.83} & \textbf{6.288} & 77.54 $\pm$ 0.19 \\
\midrule
ResNet-152 & 60.19 & 11.559 & 77.78 $\pm$ 0.43 \\
\quad + MoCE & 60.15 / \textbf{47.84} & \textbf{9.156} & 78.24 $\pm$ 0.12 \\
\bottomrule
\end{tabular}
\end{table}

\subsection{Conditional Channel Methods}

Table~\ref{tab:baselines} compares methods retrained under the same ResNet-50 protocol. MoCE and SE have essentially the same accuracy point estimate, but SE adds resources whereas MoCE removes them: MoCE reaches SE-level accuracy with $6.8$\,M fewer deployed learned parameters and $17\%$ fewer MACs. CondConv's parameter cost is consistent with our motivation that duplicating dense operators can be an inefficient route to specialization. Pick-or-Mix reaches slightly lower MACs; MoCE has the higher accuracy point estimate and comparable measured batch time. These comparisons establish the trade-off against conditional channel methods, while the distinction from static structured sparse projections is architectural rather than a claim of universal superiority.

\begin{table}[t]
\centering
\scriptsize
\setlength{\tabcolsep}{3.2pt}
\caption{Matched ResNet-50 comparison on ImageNet-1K. MoCE parameters are training/deployment counts of learned values.}
\label{tab:baselines}
\begin{tabular}{lccc}
\toprule
Method & Params (M) & MACs (G) & Top-1 (\%) \\
\midrule
Dense & 25.56 & 4.112 & 75.98 $\pm$ 0.30 \\
SE & 28.07 & 4.120 & 76.64 $\pm$ 0.19 \\
CondConv & 55.97 & 4.150 & 76.52 $\pm$ 0.24 \\
Pick-or-Mix & 25.56 & \textbf{3.178} & 76.25 $\pm$ 0.26 \\
MoCE & 25.55 / \textbf{21.26} & 3.426 & 76.71 $\pm$ 0.20 \\
\bottomrule
\end{tabular}
\end{table}

\subsection{CIFAR-100, Transfer, and EfficientViT}

Table~\ref{tab:transfer} reports CIFAR-100 trained from scratch and ImageNet-to-CIFAR transfer. MACs fall by $17.2$--$21.0\%$ throughout. The accuracy point estimate increases clearly for ResNet-50 ($+0.91$ from scratch, $+1.03$ transferring to CIFAR-100) and the margin narrows with depth, with ResNet-152 preserved within observed variation in both from-scratch CIFAR-100 and transfer to CIFAR-100. Transfer is the more informative setting for the design claim, since supports are learned on ImageNet and reused unchanged: they remain effective after fine-tuning on a different label space, which is what one expects if a channel's role is comparatively stable across inputs and tasks.

Table~\ref{tab:vit} reports EfficientViT, where the replaced layer is the second feed-forward projection rather than a bottleneck entrance. MACs fall by $18$--$22\%$ on M5/M3/M2, and all three accuracy differences are small relative to the observed run-to-run variation, so we read the result as accuracy preservation at lower computation---sufficient to establish that MoCE applies beyond ResNet-style backbones to token-wise linear projections, without claiming an accuracy gain there.

\begin{table}[t]
\centering
\scriptsize
\setlength{\tabcolsep}{2pt}
\caption{CIFAR-100 from scratch (left) and ImageNet$\rightarrow$CIFAR transfer (right); three seeds.}
\label{tab:transfer}
\begin{tabular}{lccccc}
\toprule
 & \multicolumn{2}{c}{From scratch} & \multicolumn{3}{c}{Transfer} \\
\cmidrule(lr){2-3}\cmidrule(lr){4-6}
Model & MACs & C-100 & MACs & C-10 & C-100 \\
\midrule
ResNet-50   & 1.305 & 78.44\,{\tiny$\pm$.32} & 4.11 & 97.52\,{\tiny$\pm$.10} & 85.44\,{\tiny$\pm$.20} \\
\quad + MoCE & \textbf{1.081} & 79.35\,{\tiny$\pm$.35} & \textbf{3.43} & 97.76\,{\tiny$\pm$.12} & 86.47\,{\tiny$\pm$.27} \\
\midrule
ResNet-101  & 2.520 & 78.93\,{\tiny$\pm$.32} & 7.83 & 97.80\,{\tiny$\pm$.11} & 87.50\,{\tiny$\pm$.11} \\
\quad + MoCE & \textbf{2.015} & 79.53\,{\tiny$\pm$.19} & \textbf{6.29} & 98.06\,{\tiny$\pm$.13} & 88.11\,{\tiny$\pm$.10} \\
\midrule
ResNet-152  & 3.737 & 79.40\,{\tiny$\pm$.27} & 11.56 & 97.88\,{\tiny$\pm$.06} & 88.36\,{\tiny$\pm$.16} \\
\quad + MoCE & \textbf{2.952} & 79.61\,{\tiny$\pm$.12} & \textbf{9.16} & 98.33\,{\tiny$\pm$.07} & 88.46\,{\tiny$\pm$.22} \\
\bottomrule
\end{tabular}
\end{table}

\begin{table}[t]
\centering
\scriptsize
\setlength{\tabcolsep}{4pt}
\caption{MoCE in EfficientViT on CIFAR-100 ($k{=}64$, second feed-forward projection, $s{=}2$), three seeds. Parameter counts are training-time; the deployment packing reported for ResNet is not applied here.}
\label{tab:vit}
\begin{tabular}{lccc}
\toprule
Model & Params (M) & MACs (G) & Top-1 (\%) \\
\midrule
EfficientViT-M5 & 12.13 & 0.526 & 75.56 $\pm$ 0.23 \\
\quad + MoCE    & 12.11 & \textbf{0.410} & 75.95 $\pm$ 0.17 \\
EfficientViT-M3 & 6.61  & 0.265 & 75.12 $\pm$ 0.10 \\
\quad + MoCE    & 6.60  & \textbf{0.210} & 75.20 $\pm$ 0.05 \\
EfficientViT-M2 & 3.99  & 0.204 & 74.61 $\pm$ 0.39 \\
\quad + MoCE    & 3.98  & \textbf{0.167} & 74.75 $\pm$ 0.21 \\
\bottomrule
\end{tabular}
\end{table}

\subsection{Where Input Dependence Pays}
\label{sec:adaptivity}

The upper half of Table~\ref{tab:ablation} contrasts two ways of using the input: to choose which channels an expert reads, or to choose how it combines them. All entries in this subsection are single runs on ResNet-50/CIFAR-100.

\paragraph{Input-conditioned support selection.}
The dynamic variant augments the full MoCE model: it retains the input-adaptive temperature and the residual aggregate, and replaces only the static support of Eq.~\ref{eq:support} with an input-conditioned one. Both downstream components follow the example-dependent support: the temperature is evaluated from $z_{\mathcal{S}_e(X)}$, and the residual set is recomputed per example as $\mathcal{U}(X) = \{1,\ldots,C\}\setminus\bigcup_e \mathcal{S}_e(X)$. Because a router must score every candidate channel before selecting, it reads the full pooled descriptor $z = \operatorname{GAP}(X)$ and adds a rank-$d$ factorization of the expert-by-channel offset matrix to the routing logits:
\begin{equation}
L_e(X)_c = L_{e,c} + \big\langle q_e \odot h(z),\, v_c \big\rangle,
\quad
\mathcal{S}_e(X) = \operatorname{TopK}\!\big(L_e(X), k\big),
\label{eq:dynamic}
\end{equation}
where $h(z) = \mathrm{ReLU}(W_h z + b_h) \in \mathbb{R}^{d}$ is a shared bottleneck of the descriptor, $v_c \in \mathbb{R}^{d}$ is a learned channel embedding and $q_e \in \mathbb{R}^{d}$ a learned expert embedding. The low-rank form is what keeps a per-expert, per-channel offset affordable: a separate dense network per expert would emit $(E-1)C$ values per example and be far larger than the projection it routes. The router consists of $W_h\in\mathbb{R}^{d\times C}$ and $b_h\in\mathbb{R}^{d}$, the channel embeddings $\{v_c\}_{c=1}^{C}$, and the expert embeddings $\{q_e\}_{e=1}^{E-1}$, so its parameter count per layer is $Cd + d + Cd + (E-1)d = 2Cd + Ed$ exactly; no further biases are used. With $d=32$ this is $0.96$\,M over the sixteen ResNet-50 layers. The input-conditioned logits $L_e(X)$ are used both for selection and, in place of $L_{e,i}$ in Eq.~\ref{eq:weights}, for the mixing coefficients, so the variant makes both membership and mixing weights example dependent.

During training the forward pass uses hard top-$k$ assignments and gradients are supplied by a straight-through Gumbel-softmax relaxation of the discrete selection~\citep{Jang2017Gumbel}; the input dependence comes from the router, not from the sampling noise, which supplies only the relaxation. At evaluation the variant uses deterministic hard per-example top-$k$ with no noise and no relaxation, and the reported timing is for that inference path.

In the evaluated run this variant changes accuracy by $+0.03$ points ($79.38$ vs.\ $79.35$, mean over 3 seeds), while the routing-and-gather path rises from $1.9$ to $13.5$\,ms---a factor of $7.13$. These figures time the selection and gather stages only: for MoCE the fixed indexed gather, and for the dynamic variant the router evaluation, hard per-example top-$k$, and per-example gather. They exclude temperature prediction, the channel aggregation of Eq.~\ref{eq:expert}, and the construction and aggregation of $\mathcal{U}$, which are common to both and unchanged in form. The comparison is therefore a like-for-like measurement of the path that dynamic membership alters, not a complete layer latency.

\paragraph{Input-conditioned aggregation.}
Removing the temperature gate and fixing $\tau\equiv1$ reduces accuracy by $1.19$ points. Since the static control retains fully learnable logits and the complete static function class described in Sec.~\ref{sec:temperature}, this is consistent with example-dependent concentration mattering beyond a better constant sharpness.

\paragraph{Reading the two together.}
The comparison is deliberately unequal and favors the dynamic variant, which reads the full $C$-dimensional descriptor, carries $0.96$\,M extra parameters, and makes the mixing weights example dependent as well, whereas the temperature gate reads only the $k$ descriptors on the expert's support and emits one scalar. It therefore evaluates two practical implementations rather than providing a capacity-matched isolation. Its force is that the strictly more expressive and more expensive route produces no meaningful gain in the evaluated run while forfeiting static execution, whereas removing the far more constrained route---one shared two-layer gate---reduced accuracy by $1.19$ points. These results suggest that, in the evaluated design, input dependence is more effective when applied to aggregation than to support membership. Removing the residual aggregate reduced accuracy by a further $0.74$ points.

\paragraph{How much of the dense projection is needed.}
The support-design rows deserve more attention than a routine ablation. \emph{Deterministic fixed supports} assign channels cyclically without learning:
\begin{equation}
\mathcal{S}_e^{\mathrm{cyc}} = \Big\{\, 1 + \big(((e-1)k + j) \bmod C\big) \;:\; j = 0,\ldots,k-1 \,\Big\},
\end{equation}
which gives approximately uniform channel usage and controlled overlap---exact uniformity would require $C$ to divide $(E-1)k$---with logits and temperature still trained. This reaches $78.27\%$ against the dense projection's $78.44\%$. A nonnegative convex mixture of eight channels per output, with supports assigned by a fixed rule, therefore recovers the dense projection to within $0.17$ points while removing $97\%$ of its MACs. Learning the supports adds a further $1.08$ points and overtakes the dense projection. The \emph{evaluated random assignment}---each expert draws $k$ distinct channels uniformly without replacement, independently across experts, with a single draw held fixed for the run---performs poorly at $61.30\%$. Read together, the three rows indicate that the choice of support matters and is not benign: the random assignment we evaluate degrades the representation severely, a deterministic balanced assignment nearly preserves it, and a learned one improves on the dense projection. They also suggest substantial redundancy in these projections, since eight of $C$ channels per output suffice to match the dense layer.

\subsection{Sparsity, Coverage, and Temperature}
\label{sec:sweeps}

The lower half of Table~\ref{tab:ablation} reports the three hyperparameter sweeps, also single runs. Sparsity behaves non-monotonically around $k=8$: smaller supports ($k=4$) limit expert capacity, and larger ones ($k \ge 10$) reduce specialization. Since break-even lies at $k^\star \ge 63$ for every replaced projection (Sec.~\ref{sec:cost}), $k=8$ sits far below the cost-parity point, so the binding constraint on $k$ is representational, not computational---a useful property, because it means the operating point was not chosen to satisfy a compute budget.

For the coverage weight, the default $\lambda = 5\times10^{-4}$ gives the best point estimate. Differences across the interior of the sweep are small relative to the run-to-run variation observed in the main tables and we do not read a trend into them; only the endpoints are informative, in that too weak a weight permits concentration and too strong a weight over-constrains routing. Temperature bounds behave as the reparameterization argument in Sec.~\ref{sec:temperature} predicts: a global rescaling of $L$ is absorbed, so the interval matters only as a constraint, and the widest setting $[0.1,8]$ is the worst. We adopt $[0.2,6.0]$, which spans the useful range while remaining numerically stable.

\begin{table}[t]
\centering
\scriptsize
\setlength{\tabcolsep}{3pt}
\caption{Ablations on ResNet-50; all entries are single runs. Components, support design, and temperature bounds on CIFAR-100; sparsity $k$ and coverage weight $\lambda$ on ImageNet-1K. Timings cover the inference-time routing and gather path only.}
\label{tab:ablation}
\begin{tabular}{lcc}
\toprule
Setting & Top-1 (\%) & Routing\,+\,gather \\
\midrule
\multicolumn{3}{l}{\textbf{Where input dependence is applied (CIFAR-100)}} \\
MoCE: adaptive mixing, static support & 79.35 & 1.9\,ms \\
Input-conditioned support (Eq.~\ref{eq:dynamic}) & 79.38 & 13.5\,ms \\
Static mixing ($\tau\equiv1$, $L$ learnable) & 78.16 & 1.9\,ms \\
No residual aggregate & 78.61 & -- \\
\midrule
\multicolumn{3}{l}{\textbf{Support design (CIFAR-100)}} \\
Dense $1\!\times\!1$ projection & 78.44 & -- \\
Random fixed supports (one draw) & 61.30 & -- \\
Deterministic cyclic supports & 78.27 & -- \\
Learned supports (ours) & 79.35 & -- \\
\midrule
\multicolumn{3}{l}{\textbf{Sparsity $k$ (ImageNet)}} \\
$k=4$ / $8$ / $10$ / $16$ & \multicolumn{2}{c}{76.28 / 76.71 / 75.98 / 75.73} \\
\midrule
\multicolumn{3}{l}{\textbf{Coverage weight $\lambda$ (ImageNet)}} \\
$5{\cdot}10^{-5}$ / $5{\cdot}10^{-4}$ / $5{\cdot}10^{-3}$ & \multicolumn{2}{c}{76.06 / 76.71 / 75.92} \\
$5{\cdot}10^{-2}$ / $5{\cdot}10^{-1}$ & \multicolumn{2}{c}{76.24 / 75.99} \\
\midrule
\multicolumn{3}{l}{\textbf{Temperature bounds (CIFAR-100)}} \\
$[0.5,2]$ / $[0.2,4]$ / $[0.2,6]$ & \multicolumn{2}{c}{78.86 / 79.02 / 79.35} \\
$[0.1,8]$ / unbounded & \multicolumn{2}{c}{78.49 / 79.23} \\
\bottomrule
\end{tabular}
\end{table}

\begin{table}[t]
\centering
\scriptsize
\setlength{\tabcolsep}{3pt}
\caption{FP32 batch time (ms), batch $512$, ResNet-50 eval mode; medians after warm-up with device synchronization around each measurement. PiX and MoCE medians fall within their spread; IQRs, hardware, and library versions are in the supplement.}
\label{tab:latency}
\begin{tabular}{lrrrr}
\toprule
Scope & Dense & PiX & MoCE & MoCE/Dense \\
\midrule
End-to-end & 133.12 & 127.30 & 127.00 & 0.954 \\
All replaced ops & 19.63 & 15.82 & 15.60 & 0.795 \\
Reduce 3.0 & 2.093 & 1.375 & 1.282 & 0.613 \\
Reduce 4.0 & 2.116 & 0.802 & 0.701 & 0.331 \\
\bottomrule
\end{tabular}
\end{table}

\subsection{Measured Batch Execution Time}
\label{sec:latency}

Table~\ref{tab:latency} reports FP32 batch execution time at batch size $512$ under one common evaluation setup, with mixed precision, graph compilation, and CUDA graphs disabled. MoCE and Pick-or-Mix are effectively tied end to end, and both are faster than the dense model. Across all replaced operators, MoCE reduces time from $19.63$ to $15.60$\,ms. The advantage increases in wide projections: the representative $2048\rightarrow512$ module runs at $0.331\times$ the dense time, consistent with the $k/C$ scaling of Eq.~\ref{eq:ratio} as $C$ grows from $256$ to $2048$.

The $4.6\%$ network-level gain is smaller than the MAC reduction. Section~\ref{sec:cost} offers an explanation rather than a demonstration: the sparse path still reads the input activation tensor, so an idealized arithmetic-intensity estimate places the operator on the bandwidth-dominated side of the roofline, where runtime is governed by traffic that $k$ does not reduce. The honest efficiency claim at the network level is therefore a $17$--$21\%$ MAC reduction with a $4.6\%$ end-to-end speedup, and the analysis suggests that closing the remainder is more likely to come from memory-traffic fusion than from smaller supports.

\section{Conclusion}
MoCE replaces a dense pointwise projection with channel supports that are learned during training and frozen afterwards, together with one input-dependent temperature per expert. Across the evaluated settings it preserves or improves accuracy while reducing computation, deployment storage, and measured execution time. The design finding is that input dependence is worth more for adjusting how a fixed set of channels is mixed than for changing which channels are read: a strictly more expressive input-conditioned router gained nothing measurable at $7.13\times$ the routing-and-gather cost, while removing the far cheaper temperature gate cost $1.19$ points in the same run. Two results invite follow-up: a fixed cyclic assignment of eight channels per output already matches the dense projection to within $0.2$ points, and once sparsified the layer is bounded by memory traffic rather than arithmetic.

\bibliography{main}

\end{document}